\pdfoutput=1
\documentclass[11pt]{article}

\usepackage[nohyperref,preprint]{acl}

\usepackage{xcolor}
\usepackage{tikz}
\usepackage{pgfplots}
\pgfplotsset{compat=newest}
\usetikzlibrary{positioning, patterns, fit, calc, shapes, arrows, arrows.meta}

\usepackage{times}
\usepackage{latexsym}
\usepackage{soul}

\usepackage[T1]{fontenc}
\usepackage[utf8]{inputenc}

\usepackage{microtype}

\usepackage{inconsolata}

\usepackage{graphicx}
\usepackage{subcaption}

\usepackage{amsmath} % For math

\DeclareMathOperator*{\argmax}{argmax}

\usepackage{amssymb}

\usepackage{booktabs} % For pretty tables
\usepackage{makecell}
\usepackage{multirow}
\usepackage{tcolorbox} % For the box formatting
\usepackage{placeins} % For figure management
\usepackage{enumitem}

\usepackage[backref=page,hidelinks]{hyperref}

\definecolor{border}{HTML}{596f75} 
\definecolor{background}{HTML}{f7fbfc} 

\definecolor{main}{HTML}{5989cf}    % setting main color to be used
\definecolor{sub}{HTML}{cde4ff}     % setting sub color to be used

\newtcolorbox{boxH}[1]{
    colback = sub, 
    colframe = main, 
    boxrule = 0pt, 
    arc=2mm,
    title=#1
}
\newtcolorbox{takeawayBox}[1][]{
    colframe=black, % Light teal border color
    colback=white, % Very light blue background
    boxrule=0.5mm, % Box border thickness
    arc=2mm, % Corner roundness
    left=2mm, right=2mm, top=2mm, bottom=2mm, % Padding inside the box
    width=0.45\textwidth, % Box width % Default title of the box
    #1 % Allow passing optional parameters
}

\title{When Punctuation Matters: A Large-Scale Comparison of Prompt Robustness Methods for LLMs}
\author{
 \textbf{Mikhail Seleznyov\textsuperscript{1,2}},
 \textbf{Mikhail Chaichuk\textsuperscript{1,5}},
 \textbf{Gleb Ershov \textsuperscript{3}},
 \\
 \textbf{Alexander Panchenko\textsuperscript{1,2}},
 \textbf{Elena Tutubalina\textsuperscript{1,6,7}},
 \textbf{Oleg Somov\textsuperscript{1,4}}
\\
 \textsuperscript{1}AIRI,
 \textsuperscript{2}Skoltech,
 \textsuperscript{3}Yandex,
 \textsuperscript{4}MIPT \\
 \textsuperscript{5}HSE University,
 \textsuperscript{6}Sber AI,
\textsuperscript{7}ISP RAS Research Center for Trusted AI
\\
 \small{
   \textbf{Correspondence:} \href{mailto:seleznev@airi.net}{seleznev@airi.net}, \href{mail:to:tutubalina@airi.net}{tutubalina@airi.net}, \href{mailto:somov@airi.net}{somov@airi.net}
 }
}

\begin{document}
\maketitle
\begin{abstract}

% Large Language Models (LLMs) are highly sensitive to subtle, non-semantic variations in prompt phrasing and formatting. In this work, we present a first systematic evaluation of established methods for improving prompt robustness—including Batch Calibration, Template Ensembles, and Consistency Learning—within a unified experimental framework. We benchmark these techniques across three LLM families of varying sizes, under both fine-tuned and in-context learning paradigms, across diverse prediction tasks, and under multiple types of distribution shift and compare them to baselines. Our findings offer actionable insights into the relative effectiveness of these robustness methods, enabling practitioners to make informed decisions when aiming for stable and reliable LLM performance in real-world applications.

% Mike S. May 19 8:14
Large Language Models (LLMs) are highly sensitive to subtle, non-semantic variations in prompt phrasing and formatting. In this work, we present the first systematic evaluation of 5 methods for improving prompt robustness within a unified experimental framework. 
We benchmark these techniques on 8 models from Llama, Qwen and Gemma families across 52 tasks from Natural Instructions dataset. 
Our evaluation covers robustness methods from both fine-tuned and in-context learning paradigms, and tests their generalization against multiple types of distribution shifts.
Finally, we extend our analysis to GPT-4.1 and DeepSeek V3 to assess frontier models' current robustness to format perturbations.
Our findings offer actionable insights into the relative effectiveness of these robustness methods, enabling practitioners to make informed decisions when aiming for stable and reliable LLM performance in real-world applications. Code: \urlstyle{tt}\url{https://github.com/AIRI-Institute/when-punctuation-matters}.

\end{abstract}

\section{Introduction}
\label{sec:intro}

Large Language Models (LLMs) today excel across a wide range of tasks in both in-context learning (ICL) and supervised fine-tuning (SFT) paradigms \cite{brown2020languagemodelsarefewshotlearners, gao-etal-2021-making, dong-etal-2024-survey, le2023bloom, yangetal2024unveiling, wu2024mixtureskills, mosbach2023fewshotvsincontext, yang2024iclrsupervisedknowledge, chen2024emnlpretrieveforincontextlearning}.

However, a critical yet often overlooked challenge is the high sensitivity of LLMs to prompt formatting. Many large-scale task-rich benchmarks rely on a single instruction format to evaluate all language models on a wide-range of tasks, implicitly assuming that performance is independent of prompt format \cite{hendrycks2020measuring, srivastava2023beyond}. Recent work shows that even \emph{semantically neutral} variations in prompt structure can lead to substantial changes in model predictions, often exceeding the variability introduced by model architecture or inference method~\cite{voronov2024mind, mizrahi2023state}.

% Define required colors
\definecolor{lgrey}{HTML}{EAEBEA}
\definecolor{lyellow}{HTML}{FFF3C2}
\definecolor{dgrey}{HTML}{50514F}

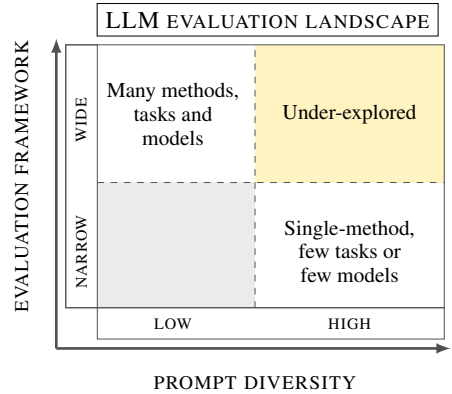
\begin{figure}[!t]
\begin{center}
\scalebox{.83}{
\begin{tikzpicture}
\begin{scope}[every node/.style={align=center, scale=0.8}]

% Lower-left.
\draw[fill=lgrey,draw=none] (1.5,1) rectangle (4,3);

% upper-left.
\node at (2.7,4.1) {\large Many methods, \\ \large tasks and \\ \large models};

% upper-right.
\draw[fill=lyellow,draw=none] (4,3) rectangle (7,5.2);
\node at (5.5,4.1) {\large Under-explored};

% lower-right.
\node at (5.5,1.9) {\large Single-method,\\ \large few tasks or \\ \large few models};

\end{scope}

% title
\node[draw] at (4.2,5.6) {\textsc{LLM evaluation landscape}};

% border and axes
\draw[draw=dgrey,thin] (1,1) rectangle (7,5.2);
\draw[draw=dgrey,thin,dashed] (4,1) -- (4,5.2);
\draw[draw=dgrey,thin,dashed] (1.5,3) -- (7,3);

% arrows
\draw[draw=dgrey, very thick, ->,>=latex] (0.85,0.35) -- (0.85,5.3);
\draw[draw=dgrey, very thick, ->,>=latex] (0.85,0.35) -- (7.1,0.35);

% x-axis
\draw[draw=dgrey, thin] (1.5,0.5) rectangle (7,1);
\node[align=center] at (2.7,0.75) {\small \textsc{low}};
\node[align=center] at (5.5,0.75) {\small \textsc{high}};
\node[align=center] at (4,-0.2) {\textsc{prompt diversity}};

% y-axis
\draw[draw=dgrey, thin] (1,1) rectangle (1.5,5.2);
\node[rotate=90, align=center] at (1.25,1.9) {\small \textsc{narrow}};
\node[rotate=90, align=center] at (1.25,4.0) {\small \textsc{wide}};
\node[rotate=90, align=center] at (0.3,3.1) {\textsc{evaluation framework}};

\end{tikzpicture}
}
\end{center}
\caption{
%It has become common in NLP to compare methods holistically, running evaluations over multiple tasks and models. However, the recently emerged field of prompt sensitivity mitigation has not yet caught up to this trend. 
Most existing robustness methods are evaluated in isolation and in disparate settings, disallowing apples-to-apples comparison.
%Most existing robustness methods are evaluated on narrow axes—either on fixed prompt templates or limited model settings. 
Our work targets the under-explored upper-right quadrant by evaluating multiple prompt robustness techniques across a wide range of prompt formats, LLM families, learning paradigms, and distribution shifts under a unified framework.
}
\label{fig:challenge}
\end{figure}

\textbf{Prompt format} (e.g. spacing, capitalization, punctuation)
can dramatically influence performance, leading to inconsistent or unreliable outputs~\cite{zhao2021calibrate, min-etal-2022-noisy}. This phenomenon, known as \textbf{prompt sensitivity}, can be mitigated using specialized \textbf{robustness methods}.

A number of robustness techniques have been proposed to address this issue, including Template Ensembling~\cite{voronov-etal-2024-mind}, Sensitivity-Aware Decoding~\cite{lu-etal-2024-prompts}, Batch Calibration~\cite{zhou2024batchcalibration}, and Consistency Learning~\cite{qiang-etal-2024-prompt}. However, these methods have primarily been evaluated in isolation, making it difficult for practitioners to assess their relative strengths or determine which method is best suited to a given scenario.

Our work addresses this gap through a comprehensive, systematic evaluation of prompt robustness techniques under a unified experimental framework. Specifically, we benchmark four widely cited robustness methods against standard few-shot prompting and fine-tuning with prompt format augmentation as baselines.

We conduct experiments using a representative subset of 52 tasks from the well-known Natural Instructions dataset, covering domains such as mathematics, logic, and text comprehension. As backbone models, we evaluate three modern LLM families~--- \textsc{Gemma} \cite{gemma2024kaggle}, \textsc{LLaMA} \cite{dubey2024llama3herd}, and \textsc{Qwen} \cite{yang2025qwen25technicalreport}~--- with sizes from 1.5B to 9B parameters. We additionally include closed-source models to study format sensitivity at scale. Within our framework, we answer the following research questions:

%[label=\arabic*., labelsep=0.5em, leftmargin=*]
\begin{enumerate}
    \item[\textbf{RQ1:}] How do existing robustness methods compare in effectiveness across various models?
    \item[\textbf{RQ2:}] How distribution shifts affect the effectiveness of SFT-based and ICL-based methods?
    \item[\textbf{RQ3:}] How does greedy decoding affect robustness compared to choosing highest-probability answer option?
    \item[\textbf{RQ4:}] How sensitive are frontier models  to format perturbations, and what methods can be applied in black-box setting to improve their robustness?
\end{enumerate}

To the best of our knowledge, this is the first study to offer a side-by-side comparison of multiple prompt robustness methods under a unified, large-scale evaluation protocol spanning diverse prompt formats, model families, learning paradigms, distribution shifts and inference strategies. By bridging previously disconnected lines of work, our findings provide actionable insights for both practitioners and researchers interested in building more stable and reliable LLM-based systems. We also release our code to encourage systematic evaluation in the field of prompt sensitivity mitigation.

\section{Related Work}

Recent work has highlighted the sensitivity of language models to subtle prompt variations, but current research remains fragmented~\cite{zhuo-etal-2024-prosa, pei-etal-2025-selfprompt}.
%across adversarial robustness, calibration, and prompt structure analysis.
Adversarial-focused studies~\cite{zhu2024promptrobust, zou2023universalandtransferableadversarialattacks} expose vulnerabilities to malicious or perturbed prompts, emphasizing safety but targeting directed threat models rather than benign formatting inconsistencies.

Other works propose robustness-enhancing methods such as Consistency Learning~\cite{qiang-etal-2024-prompt}, Batch Calibration~\cite{zhou2024batchcalibration}, and Template Ensembles~\cite{voronov2024mind}, which improve stability either during training or inference. However, these approaches are evaluated in isolation, making it difficult to assess their relative effectiveness.

Complementary studies~\cite{lu-etal-2024-prompts, zhao2021calibrate, sclar2024quantifyinglanguagemodelssensitivity} analyze prompt components and formatting artifacts, showing that even innocuous design choices (e.g., whitespace, punctuation) can introduce large performance shifts. This further underscores the need for unified, standardized evaluation protocols.

In summary, while prior research has addressed different aspects of prompt sensitivity, there is a lack of systematic, comparative evaluation across tasks, models, and learning paradigms. Our work fills this gap by benchmarking four robustness methods under a unified framework across 52 diverse tasks, multiple LLM families and distribution shift scenarios, resulting in  actionable takeaways for practitioners.

\section{Experimental Setup}

To answer our research questions, we use a subset of Natural Instructions with a parametrized set of formats (Section \ref{sec:experimental_setup_data_format}) and implement the methods from Section \ref{sec:experimental_setup_methods}. We evaluate performance and robustness using the metrics defined in Section \ref{sec:metrics_and_inference}.

%In this section we describe what dataset we use for evaluations, how do we define format, the compared methods, as well as metrics and inference strategies (greedy decoding or selecting highest-probability answer option).

\begin{table*}[!ht]
    \centering
    \begin{tabular}{cccccc}\toprule
        \makecell{Descriptor\\transformation} & Separator & Space & \makecell{Text \& option\\separator} & \makecell{Option item style} & \makecell{Option item wrapper} \\ \midrule
        \makecell{\texttt{.title()} \\ \texttt{.uppercase()} \\ \texttt{.lowercase()}} & \makecell{\texttt{`: '} \\ \texttt{`- '} \\ \texttt{`\textbackslash n'} } & \makecell{\texttt{` '} \\ \texttt{`\textbackslash n'} \\ \texttt{`; \textbackslash n'} } & \makecell{\texttt{` '} \\ \texttt{`\textbackslash t'} \\ \texttt{`'}} & \makecell{\texttt{A, B, C, ...} \\ \texttt{1, 2, 3, ...} \\ \texttt{I, II, III, ... }} & \makecell{\texttt{\{\})} \\ \texttt{\{\}.} \\ \texttt{[\{\}]}}
        \\ \bottomrule
    \end{tabular}
    \caption{Format components, with some example values. Descriptor transformation correspond to Python command making first character upper case (title), all letters upper case (uppercase) or all letters lower case (lowercase).  For option item wrapper, $\{\}$ is used as a placeholder for option item (e.g. A or 1).}
    \label{tab:format_structure}
\end{table*}

\subsection{Data \& Format}
\label{sec:experimental_setup_data_format}

We use a subset of 52 tasks from Natural Instructions \cite{wang-etal-2022-super} with diverse human-written formats and instructions, comprising 19 multiple-choice tasks and 33 classification tasks with {2, 3 or 4} answer options. Given that there are more than 1600 tasks we select relevant and socially impact tasks following \citet{sclar2024quantifyinglanguagemodelssensitivity} task selection criteria (refer to Appendix \ref{sec:selection} for details). Resulting tasks cover math and logic problems, text comprehension, detection of harassment and racial stereotypes. To evaluate the performance, we use a subset of $1000$ random examples from each task.

\paragraph{Prompt format.} We consider 6 types of format components, following \citet{sclar2024quantifyinglanguagemodelssensitivity}. They are listed in Table \ref{tab:format_structure}.
% they are listed in Table \ref{tab:format_structure} in Appendix. 
For each component there are between 4 and 16 possible values. To construct a format, we select a specific value of each component. For example, default prompt for a task could be structured as following: \textbf{$\texttt{{question:\{\}A)\{\}B)\{\}answer:\{\}}}$}, where \textbf{\texttt{\{\}}} denotes the placeholders for the task instruction and multiple-choice answer options (following \cite{wang-etal-2022-super} formatting).
Then choosing the first values from first row of Table \ref{tab:format_structure} to modify original prompt design results in 
\begin{equation*}
    \texttt{Question: \{\} A) \{\} B) \{\} Answer: \{\}}
\end{equation*}
whereas taking values from second row forms another prompt design:
\begin{equation*}
    \texttt{QUESTION- \{\}\textbackslash n1.\textbackslash t\{\}\textbackslash n2.\textbackslash t\{\}\textbackslash nANSWER- \{\}}
\end{equation*}

For some tasks there are no multiple choice options~--- in this case the format is defined only by descriptor transformation, separator and space.
Complete list of format components is available in Appendix \ref{sec:all_format_components}.

\subsection{Methods}
\label{sec:experimental_setup_methods}

We consider five representative approaches for improving robustness to prompt formatting. These span both ICL and SFT paradigms. Below, we briefly describe each method.

\paragraph{Few-shot (FS).}
As a baseline, we use a standard 2-shot prompting strategy. 
Since the selection and order of demonstration examples can significantly influence results \cite{lu-etal-2022-fantastically}, we fix the in-context examples and their order across all models and test samples.
Demonstration examples are also formatted in the same way as the test sample, hinting to the model that formatting should not influence the prediction.

\paragraph{Batch Calibration (BC).}
Batch Calibration~\cite{zhou2024batchcalibration} is a post-hoc correction technique. It estimates contextual bias across a batch and adjusts predicted log-probabilities by subtracting the bias. While simple and efficient, it is limited to classification tasks.

\paragraph{Template Ensembles (TE).}
Template Ensembling~\cite{voronov-etal-2024-mind} improves robustness by averaging predicted class probabilities across $N$ prompt formats. This reduces variance caused by formatting changes, but increases inference cost linearly with $N$.

\paragraph{Sensitivity-Aware Decoding (SAD).}
This approach is inspired by~\citet{lu-etal-2024-prompts}. It penalizes predictions that are sensitive to synthetic input perturbations. In our implementation, we use random token substitutions to estimate sensitivity. This approach helps to stabilize outputs but requires multiple forward passes per input.

\paragraph{LoRA with format augmentations (LoRA).}
We apply parameter-efficient fine-tuning (PEFT) using LoRA on an instruction-following dataset augmented with formatting variations. This method exposes the model to diverse prompt styles during training with the aim of mitigating spurious correlations between answers and format components.

\paragraph{LoRA with consistency loss (LoRA-JS).}
Following~\citet{qiang2024consistency}, we add a Jensen-Shannon consistency loss between outputs of different prompt variants, encouraging format-invariant predictions. The total loss combines standard cross-entropy with a divergence term. 

% \medskip
Full implementation details for each method are provided in Appendices~\ref{sec:full_methods_description}, \ref{sec:hyperparameters}.

\begin{figure*}[ht!]
    \centering
    \includegraphics[width=\linewidth]{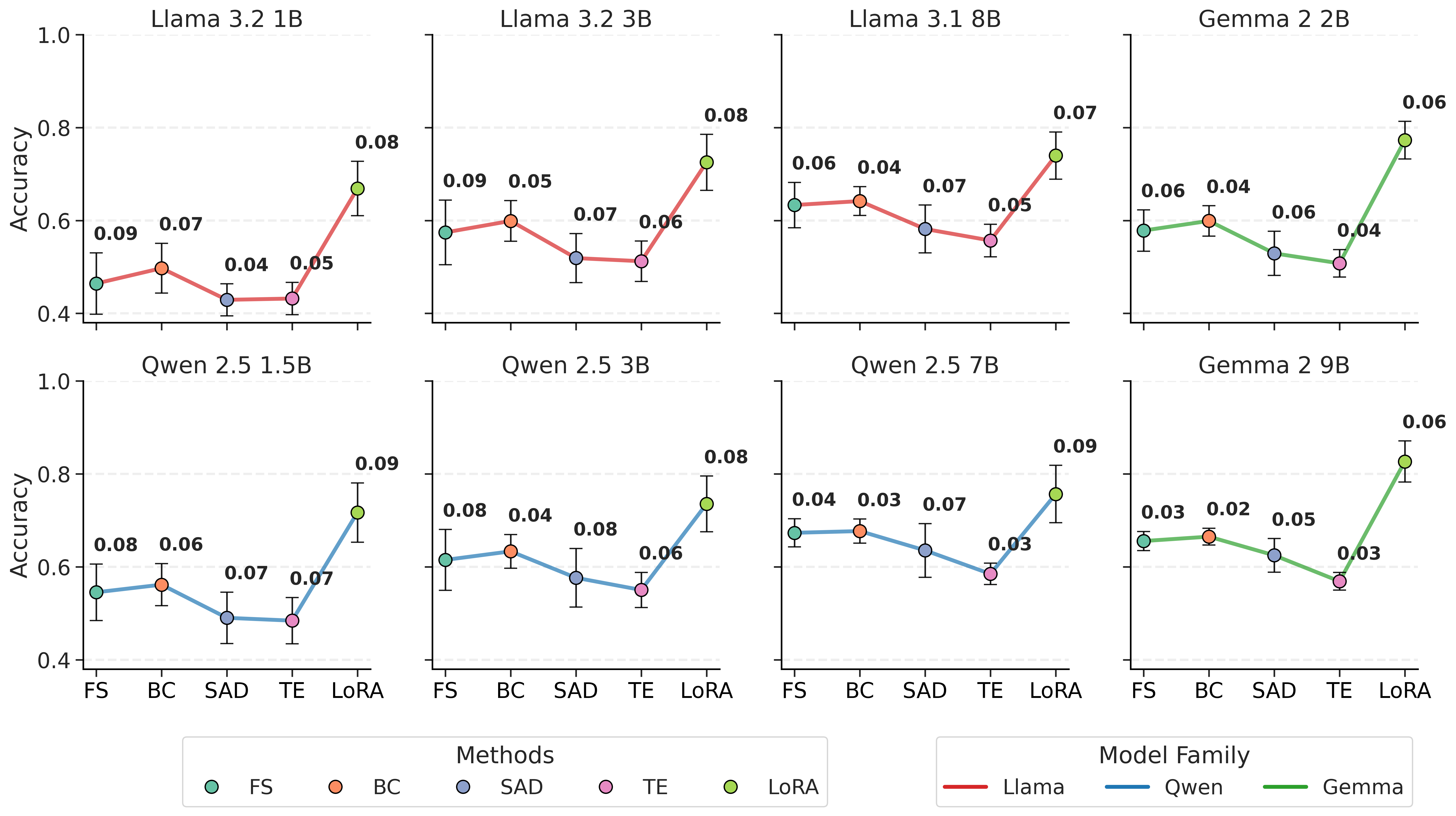}
    \caption{Comparing format sensitivity mitigation methods in terms of their effect on accuracy and standard deviation over prompt formats. To aggregate accuracy, we first compute median accuracy over formats for each task, and then average over 52 tasks. Error bars are $2\;\times$ (standard deviation over formats, averaged across tasks).}
    \label{fig:rq1_base_plot}
\end{figure*}

\subsection{Metrics \& Inference Approach}
\label{sec:metrics_and_inference}

\paragraph{Evaluation Metrics.} \label{sec:evaluation_metrics}
To evaluate model performance, we use \textit{accuracy} as the primary metric. To assess sensitivity to prompt formatting, we report two measures: \textit{spread} and \textit{standard deviation}. Spread is defined as the difference between the maximum and minimum accuracy across a set of prompt formats~\cite{sclar2024quantifyinglanguagemodelssensitivity}, providing a simple measure of output variability due to prompt variation. We also consider a class-imbalance setting. Since accuracy is often misleading on imbalanced tasks, we report \textit{Matthews correlation coefficient (MCC)}.
We choose MCC instead of F1 score since the latter puts emphasis on the positive class and may change dramatically after a permutation of classes. MCC treats classes symmetrically and accounts for all four confusion matrix components, including true negatives. Since our tasks usually do not have a distinguished positive class, MCC is better suited for our evaluations.
%\footnote{We report MCC instead of F1 score, as it more appropriately handles classification tasks where the distinction between positive and negative classes is not predefined. Unlike F1, MCC accounts for all four confusion matrix components—including true negatives—making it better suited for our evaluation.} 

\paragraph{Inference Strategies.}
To obtain answers from the language model in multiple-choice and classification tasks, we use two common inference strategies: \textit{greedy decoding} and \textit{probability ranking}. Greedy decoding generates the answer token-by-token, selecting the most likely token at each step. The result string is normalized and evaluated to gold answer with exact match. In contrast, probability ranking computes the probability of each answer option, and selects the highest-ranked one. 
%Since all answer options are known in advance, this method is implemented using teacher forcing, conditioning the model on the full sequence of each option.

\section{Results}

Our experiments address the four research questions outlined in Section \ref{sec:intro}: methods comparison in default setting (\textbf{RQ1}, Section \ref{sec:results_rq1}), effect of distribution shifts on fine-tuning and ICL-based methods (\textbf{RQ2}, Section \ref{sec:results_rq2_shifts}), effect of inference strategy (\textbf{RQ3}, Section \ref{sec:results_rq3}) and evaluations of frontier models (\textbf{RQ4}, Section \ref{sec:results_rq4_frontier}).
\subsection{Robustness Method Comparison}
\label{sec:results_rq1}

\textbf{RQ1: How existing robustness methods compare in effectiveness across different LLM families and sizes?}
%Within this research question, our primary goal is to compare
To answer \textbf{RQ1}, we apply all methods under the default conditions (without a distribution shift) to evaluate their impact on accuracy and robustness.
%Spread across different prompt formats is a primary measure for evaluating robustness. -- explained in Metrics & Inference
We assess accuracy to check whether robustness methods negatively affect performance. Figure \ref{fig:rq1_base_plot} plots accuracy of each method, averaged across 52 tasks, along with standard deviation over formats\footnote{We also tested LoRA with consistency loss. However, the results were less favorable compared to the other methods, especially LoRA with format augmentations. Some of these findings are available in Appendix \ref{sec:appendix_consistency_loss}.}.

\begin{figure}
    \centering
    \includegraphics[width=\linewidth]{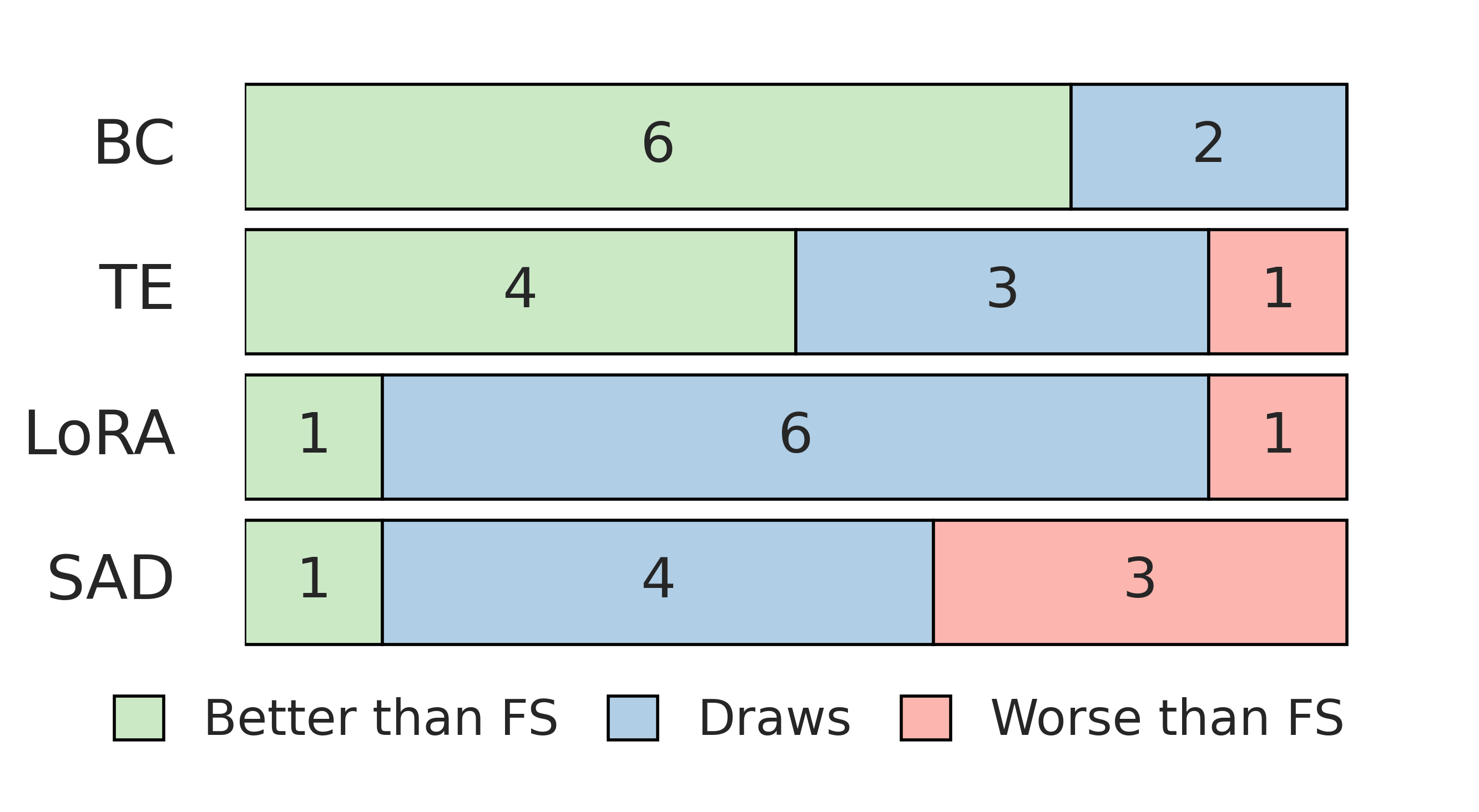}
    \caption{Comparing format sensitivity mitigation methods against regular few-shot in terms of spread on 8 language models. Method wins against few-shot for a given model if it has statistically significantly lower spread than few-shot on 52 tasks.}
\label{fig:method_baseline_battles_plot}
\end{figure}

To compare methods' effectiveness in improving robustness, we use the following procedure. For a fixed model $M_0$ (e.g. Llama 3.1 8B) and each task $t$ we estimate the spreads of baseline few-shot approach and the competing method $X$ over 10 formats, where $X \in \left\{\text{BC}, \text{TE}, \text{SAD}, \text{LoRA}\right\}$. Then, we consider differences 
\begin{gather}
    \label{eq:stattest}
    \nonumber \text{SpreadDiffs}_{M_0, X} = \\
    \left\{ \text{spread(FS)}_{t, M_0} -\text{spread(X)}_{t, M_0} \right\} \\ \nonumber \forall t \in \mathbf{T}, |\mathbf{T}| = 52,
\end{gather}
where $\mathbf{T}$ is our selected 52 tasks from Natural Instructions.
Finally, we run Student's t-test with $H_0$:~mean of $\text{SpreadDiffs}_{M_0, X}$ is equal to zero. If $H_0$ is rejected, the method with lower mean spread is considered more robust on model $M_0$. Otherwise, $X$ ties with few-shot. Such tests are run for each of 8 open-source models we evaluate, and the results are shown in Figure \ref{fig:method_baseline_battles_plot}.

\begin{figure*}
    % \centering
    \includegraphics[width=\linewidth]{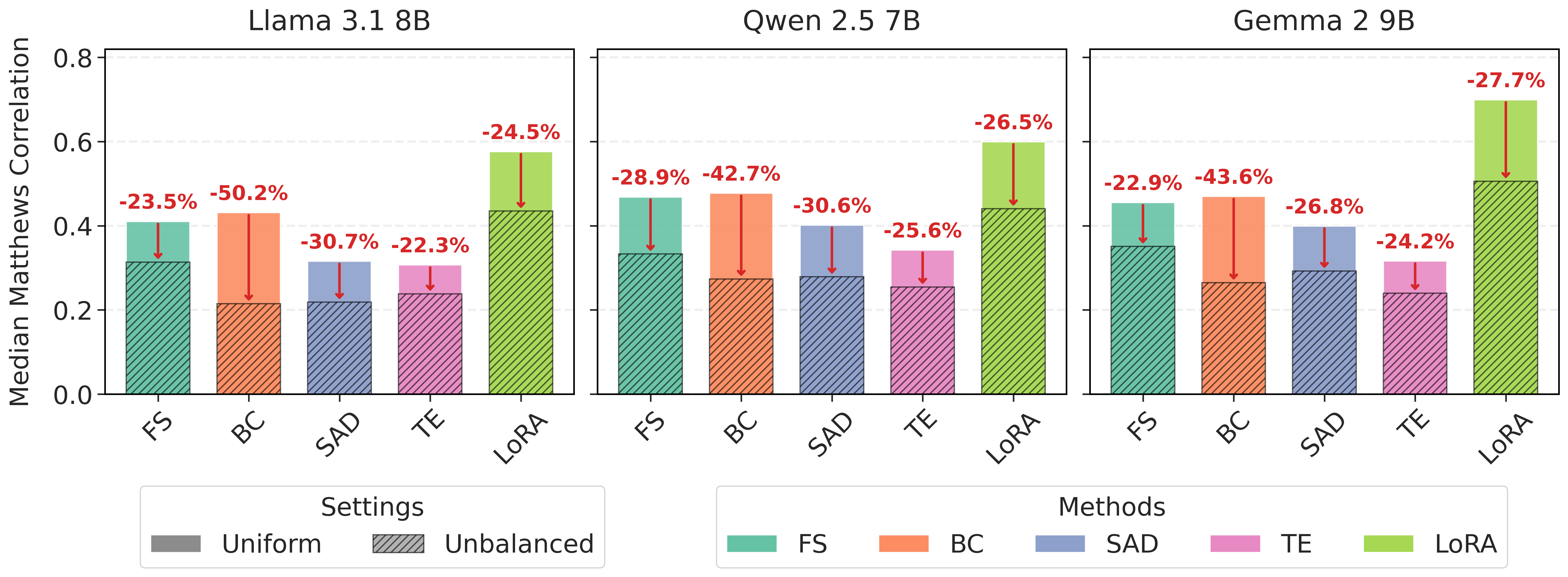}
    \caption{Median Matthews Correlation of robustness methods in uniform vs. unbalanced settings for LLaMA 3.1 8B, Qwen 2.5 7B, and Gemma 2 9B. Red values indicate the drop in performance under the unbalanced setting relative to the uniform case.}
    \label{fig:bc_failure_half}
\end{figure*}

\paragraph{Batch Calibration improves both accuracy and robustness across the board.} From Figure \ref{fig:rq1_base_plot} we can see that Batch Calibration achieves higher average accuracy compared to few-shot for all 8 open-source models. Meanwhile, Figure \ref{fig:method_baseline_battles_plot} shows that BC delivers statistically significant reduction of spread for 6/8 models.
Since calibration methods do not require training data and have near-zero inference time overhead, these strong results put Batch Calibration as a clear leader in terms of format robustness enhancement in absence of distribution shifts.

\paragraph{Template Ensembles improve robustness at cost of reducing accuracy.} Ensembling method proposed by \citet{voronov2024mind} also reduces spread, with statistically significant reductions for 4/8 models. However, it results in lower accuracy compared to few-shot baseline. To investigate the causes of the drop in performance, we inspected predictions of individual ensemble members. It turned out that for one format in the ensemble the accuracy sometimes underperforms, affecting average probabilities.
This aligns with the original findings of \citet{voronov2024mind}, which note that a single suboptimal template may make the ensemble perform noticeably worse.
Together, it suggests that logit averaging is sometimes a brittle strategy, sensitive to outliers.

\paragraph{LoRA with augmentations enhances accuracy, but struggles to consistently improve robustness.} On Figure \ref{fig:rq1_base_plot} we can see that LoRA with augmentations achieves much higher average accuracy compared to ICL-based approaches. This is expected, since LoRA is the only SFT-based method on the plot, and has access to training labels. Perhaps more surprisingly, augmentations have almost no impact on robustness: Figure \ref{fig:method_baseline_battles_plot} shows that LoRA improves spread compared to few-shot only on a single model out of 8, with 6 ties and 1 loss.
\medskip

% \begin{takeawayBox}
\textbf{Takeaways:}
    \begin{enumerate}[topsep=2pt]
        \item In absence of distribution shifts, calibration-based approach shows promise in improving robustness to prompt formats due to its ability to significantly reduce spread, positive effect on accuracy and low overhead.
        \item While naive parameter-efficient finetuning  with augmentations significantly improves accuracy, it turns ineffective in mitigating sensitivity to format changes.
        \item Probability averaging strategy used in Template Ensemble helps to reduce spread, but may suffer from sensitivity to especially poor-performing formats. 
        %In Section \ref{sec:results_rq3} with greedy decoding we explore majority voting as an alternative more stable  aggregation strategy.
        %\todo{Should we keep last sentence? Formally we can't compare aggregation strategies, since probability averaging is only used with probability ranking, and majority voting is only used with greedy decoding. However, majority voting is indeed be more stable to outliers, simply because it is based on mode rather than mean.}
    \end{enumerate}
% \end{takeawayBox}

\begin{figure}[t]
  \centering    
  {\includegraphics[width=0.5\textwidth]{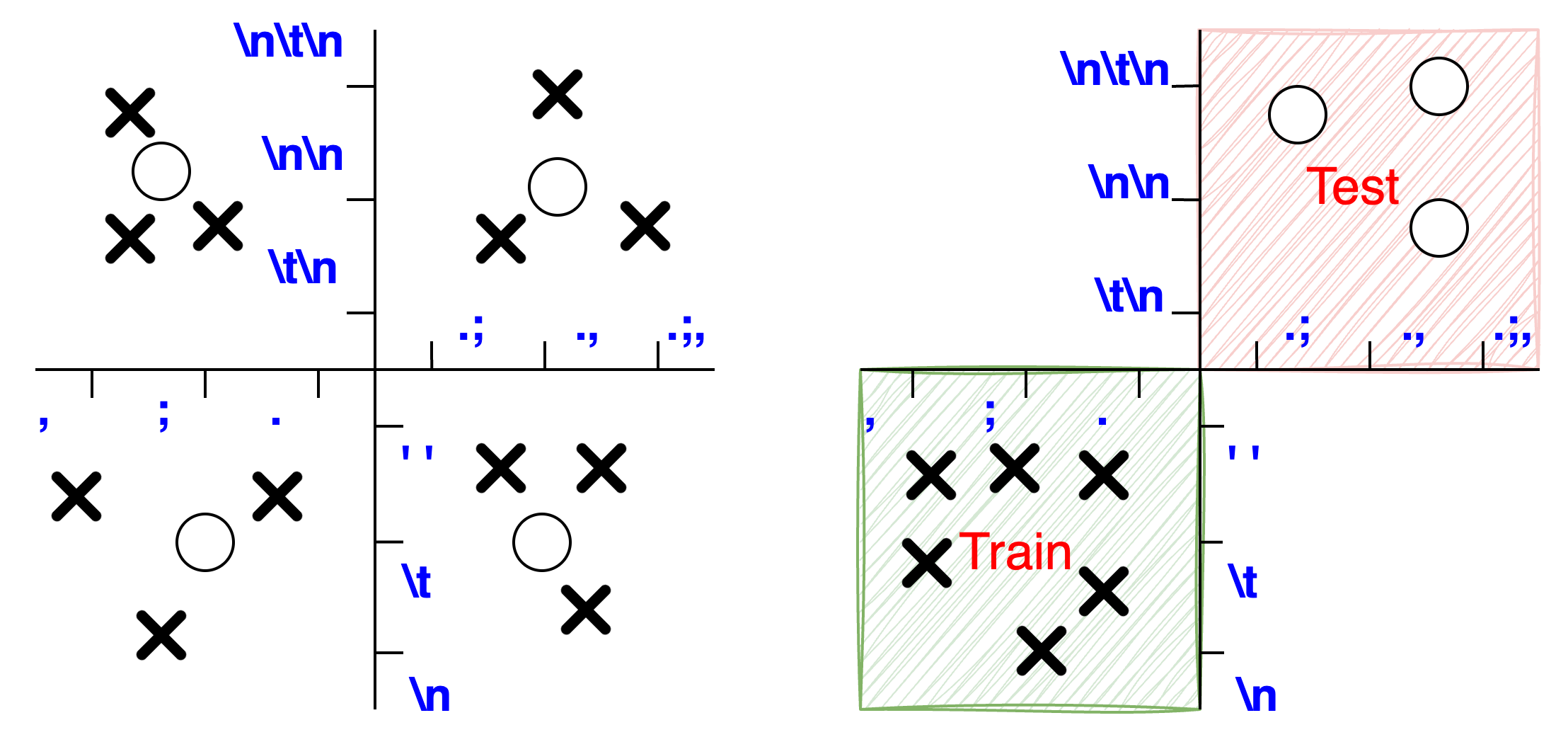}}
  \caption{Without distribution shift (left), train and test formats are sampled uniformly. Under the compositional distribution shift (right), the test set contains novel combinations of known components, requiring systematic generalization. Cross ($\times$) stands for train samples, circle ($\circ$) for test.}
  \label{fig:uni_vs_comp_scenario}
\end{figure}

% \begin{table}
% \centering
% \small
%     \begin{tabular}{llllll}
%     \toprule
%     Method & BC & FS & LoRA & SAD & TE \\
%     % Setting &  &  &  &  &  \\
%     \midrule
%     % Uniform & $2.55$ & $2.91$ & $1.71$ & $4.26$ & $3.52$ \\
%     Uniform & $2.6$ & $2.9$ & $1.7$ & $4.3$ & $3.5$ \\
%     % Unbalanced & $3.22$ & $2.69$ & $1.74$ & $4.03$ & $3.26$ \\
%     % Unbalanced & $3.22_{\textcolor{red}{(\scriptstyle+0.67)}}$ &  $2.69_{\textcolor{green}{(\scriptstyle-0.22)}}$ & $1.74_{\textcolor{red}{(\scriptstyle+0.03)}}$ & $4.03_{\textcolor{green}{(\scriptstyle-0.23)}}$ & $3.26_{\textcolor{green}{(\scriptstyle-0.26)}}$ \\
%     Unbalanced & $3.2_{\textcolor{red}{\scriptstyle+0.6}}$ &  $2.7_{\textcolor{teal}{\scriptstyle-0.2}}$ & $1.7$ & $4.0_{\textcolor{teal}{\scriptstyle-0.3}}$ & $3.3_{\textcolor{teal}{\scriptstyle-0.2}}$ \\
%     \bottomrule
%     \end{tabular}
%     \caption{Average rankings of methods across models by Matthews correlation coefficient (1 is best). Rankings are averaged across models and tasks. \label{tab:method_rankings}}
% \end{table}

\begin{table}
\centering
\small
\begin{tabular*}{\linewidth}{@{\hskip 2pt}l@{\hskip 6pt}l@{\hskip 6pt}l@{\hskip 6pt}l@{\hskip 6pt}l@{\hskip 6pt}l@{\hskip 2pt}}
\toprule
Method & BC & FS & LoRA & SAD & TE \\
\midrule
Default & $2.6$ & $2.9$ & $1.7$ & $4.3$ & $3.5$ \\
Unbalanced & $3.2_{\textcolor{red}{\scriptstyle+0.6}}$ &  $2.7_{\textcolor{teal}{\scriptstyle-0.2}}$ & $1.7$ & $4.0_{\textcolor{teal}{\scriptstyle-0.3}}$ & $3.3_{\textcolor{teal}{\scriptstyle-0.2}}$ \\
\bottomrule
\end{tabular*}
\caption{Rankings of methods across models by Matthews correlation coefficient (1 is best). Rankings are averaged across models and tasks. \label{tab:method_rankings}}
\end{table}

% \begin{table}
% \centering
% \small
% \begin{tabular*}{\linewidth}{@{\hskip 2pt}l@{\hskip 6pt}c@{\hskip 6pt}c@{\hskip 6pt}c@{\hskip 6pt}c@{\hskip 6pt}c@{\hskip 2pt}}
% \toprule
% Method & BC & FS & LoRA & SAD & TE \\
% \midrule
% Uniform & $2.6$ & $2.9$ & $1.7$ & $4.3$ & $3.5$ \\
% Unbalanced & $3.2_{\textcolor{red}{\scriptstyle+0.6}}$ &  $2.7_{\textcolor{teal}{\scriptstyle-0.2}}$ & $1.7$ & $4.0_{\textcolor{teal}{\scriptstyle-0.3}}$ & $3.3_{\textcolor{teal}{\scriptstyle-0.2}}$ \\
% \bottomrule
% \end{tabular*}
% \caption{Average rankings of methods across models by Matthews correlation coefficient (1 is best). Rankings are averaged across models and tasks. \label{tab:method_rankings}}
% \end{table}

\subsection{Impact of Distribution Shifts}
\label{sec:results_rq2_shifts}

\textbf{RQ2: How do distribution shifts affect the effectiveness of SFT-based and ICL-based methods?}
As we see in Section \ref{sec:results_rq1}, Batch Calibration helps to improve robustness while LoRA significantly stands out in terms of accuracy. To answer \textbf{RQ2}, we zoom in and inspect these methods in more detail to understand their limitations.

\paragraph{Covariate shift (class imbalance).} 

In Batch Calibration, predicted probabilities of each class are adjusted by subtracting mean probability of this class over batch. Naturally, predicted probabilities for classes that occur more often are adjusted more. They in turn are less often selected by argmax, which leads to a more uniform predictive distribution. Thus, Batch Calibration implicitly assumes more uniform class distribution compared to baseline few-shot approach. While reasonable in balanced tasks, this assumption may lead to calibration errors under class imbalance. To investigate this, we construct an artificial imbalanced dataset by downsampling each of our 52 tasks such that the most frequent class constitutes 90\% of the examples, with the remaining classes evenly splitting the remaining 10\%.

%A closer examination of the Batch Calibration formulation reveals that it implicitly assumes a uniform prior over classes in the multiple-choice setting~---that is, each label is treated as equally likely in expectation.

In Figure~\ref{fig:bc_failure_half}, we observe that all methods are affected by the unbalanced setting, but Batch Calibration suffers the most due to the model’s inductive bias toward a uniform class distribution. Table \ref{tab:ranking_vs_generation} confirms this finding: Batch Calibration exhibits the largest change in average ranking. LoRA-fine-tuned models also degrade, as they were trained assuming a uniform class distribution and thus are also subject to covariate shift.

\paragraph{Compositional and cross-domain shifts.} 
To evaluate the robustness of the LoRA method to distribution shifts, we consider two scenarios: \textbf{compositional} and \textbf{cross-domain}.

\textbf{Compositional shift.} Inspired by the notion of systematic compositionality~\cite{hupkes2020compositionality}, this setting tests the model’s ability to generalize by recombining known elements in novel ways. In default scenario in Section \ref{sec:results_rq1}, train and test formats are sampled uniformly. However, under compositional shift test formats contain new combinations of previously seen format components.
%The training data mirrors the task distribution used in the Uniform scenario, while the test data features systematically rearranged prompt components. 
%This allows us to assess whether the model has learned representations that support generalization to unseen compositions of prompt format components. 
An illustration of compositional train/test format split is shown in Figure~\ref{fig:uni_vs_comp_scenario}.

\begin{figure}[t]
    \centering
    \includegraphics[width=\linewidth]{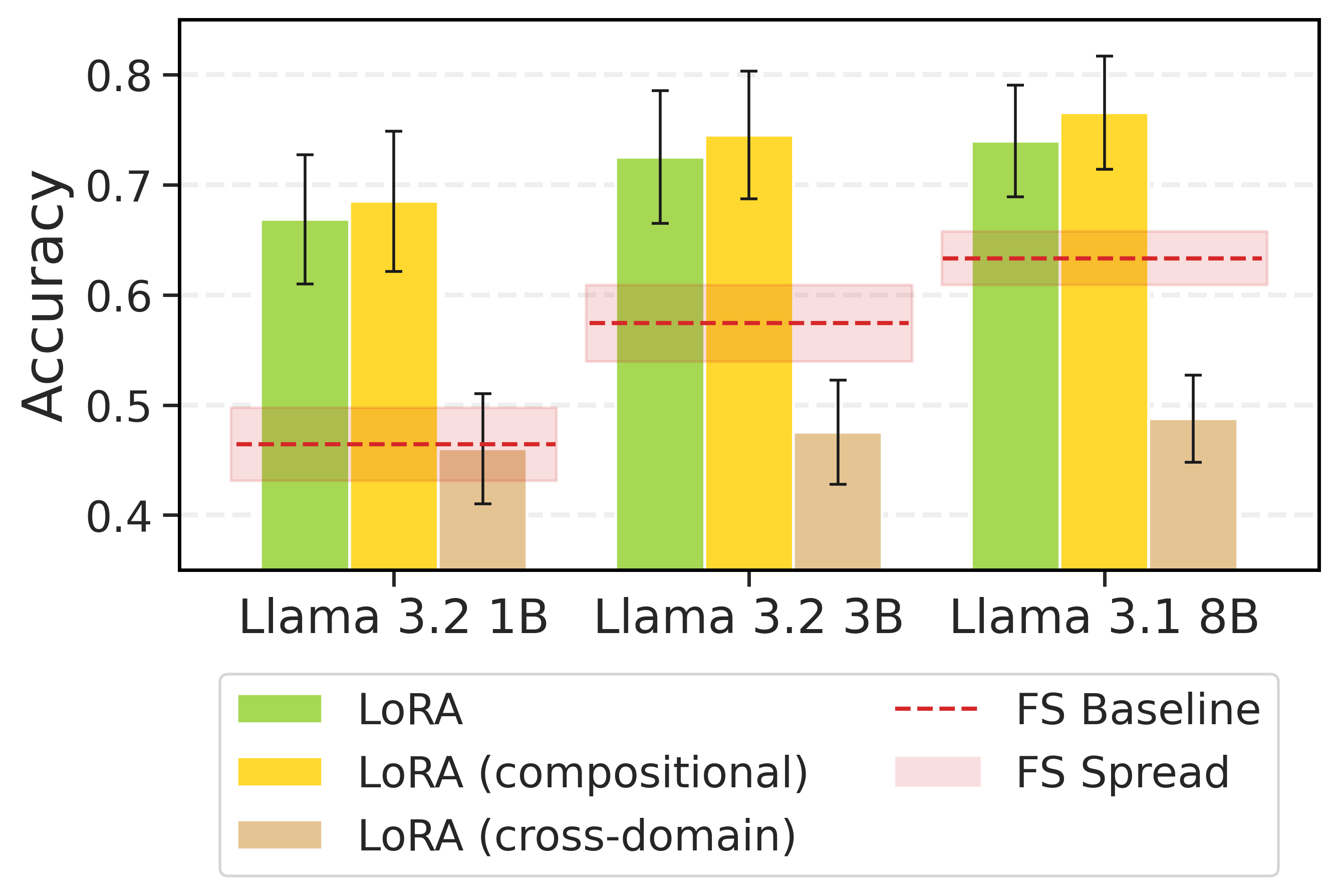}
    \caption{LoRA method under distribution shifts. To aggregate accuracy, we first compute median accuracy over formats for each task, and then average over 52 tasks. Error bars are $2\;\times$ (standard deviation over formats, averaged over tasks).}
    \label{fig:lora_comparison}
\end{figure}

\textbf{Cross-domain shift.} To evaluate robustness to domain changes, this setting uses training data from an external dataset (see Appendix \ref{par:lora_ft_details}). The prompt formats remain uniformly distributed during both training and testing like in Section \ref{sec:results_rq1}. This setup probes the model's and method's ability to disentangle semantics from format and generalize beyond the training domain.

\begin{table}
\small 
\centering
\begin{tabular}{lll}
\toprule
\makecell{Inference\\Strategy} & \makecell{Greedy\\Decoding} & \makecell{Probability\\Ranking} \\
\midrule
Gemma 2 2B & $0.48 \pm \textbf{0.28}$ & $0.58 \pm 0.06$ \\
Gemma 2 9B & $0.55 \pm \textbf{0.32}$ & $0.66 \pm 0.03$ \\\midrule
Llama 3.1 8B & $0.63 \pm \textbf{0.11}$ & $0.63 \pm 0.06$ \\
Llama 3.2 1B & $0.46 \pm \textbf{0.11}$ & $0.46 \pm 0.09$ \\
Llama 3.2 3B & $0.56 \pm \textbf{0.13}$ & $0.57 \pm 0.09$ \\\midrule
Qwen 2.5 1.5B & $0.49 \pm \textbf{0.12}$ & $0.55 \pm 0.08$ \\
Qwen 2.5 3B & $0.59 \pm \textbf{0.12}$ & $0.61 \pm 0.08$ \\
Qwen 2.5 7B & $0.63 \pm \textbf{0.14}$ & $0.67 \pm 0.04$ \\
\bottomrule
\end{tabular}
\caption{Comparison of inference strategies: greedy decoding vs. probability ranking. To aggregate accuracy, we first compute median accuracy and standard deviation over formats for each task, and then average over 52 tasks. We report averaged median accuracy $\pm \, 2\times$ averaged std. Higher standard deviation is in bold. \label{tab:ranking_vs_generation}}
\end{table}
\begin{figure*}[ht!]
    \subfloat[DeepSeek V3]{\includegraphics[width=0.5\textwidth]{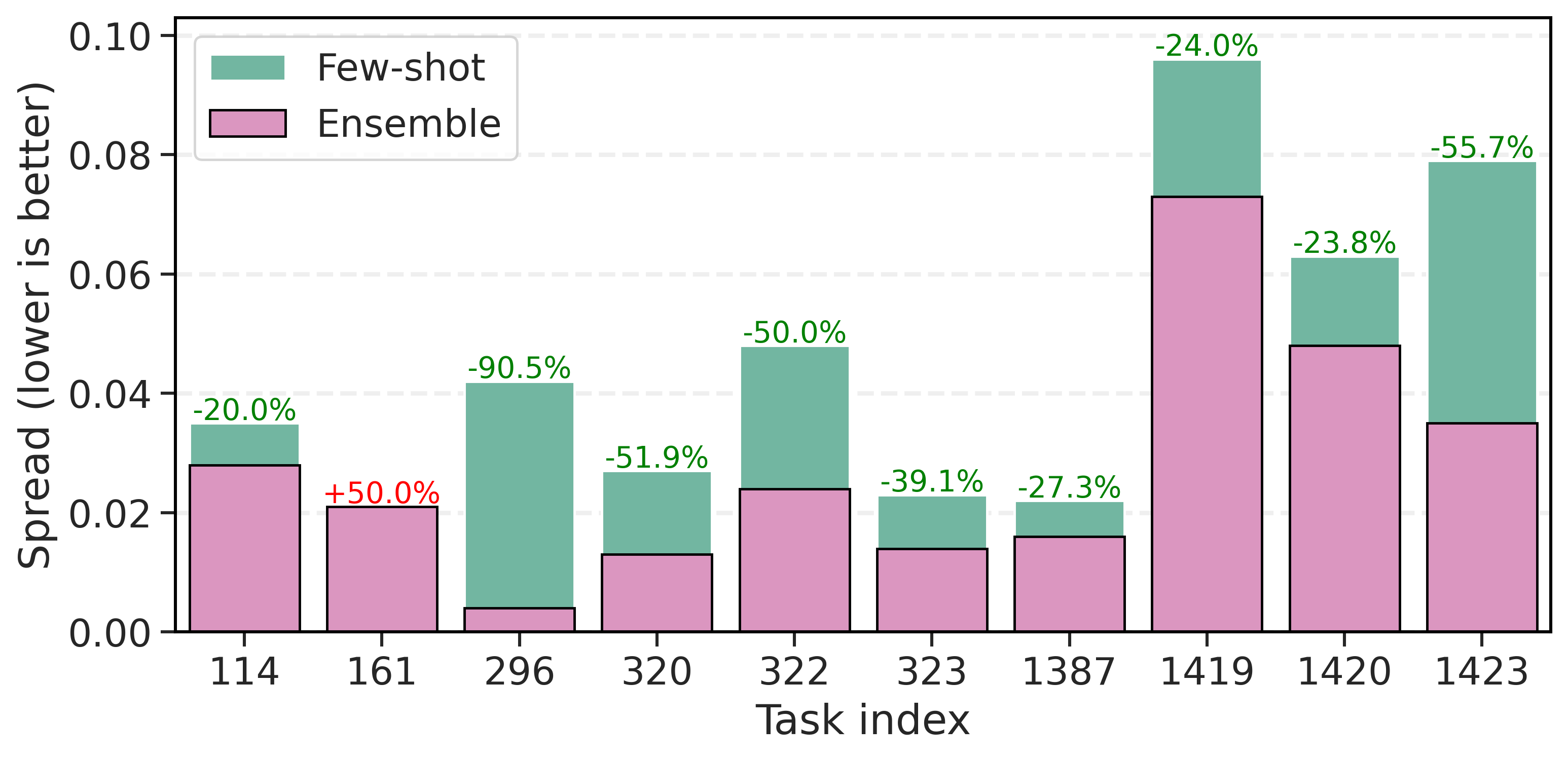}}
    \subfloat[GPT-4.1]{\includegraphics[width=0.5\textwidth]{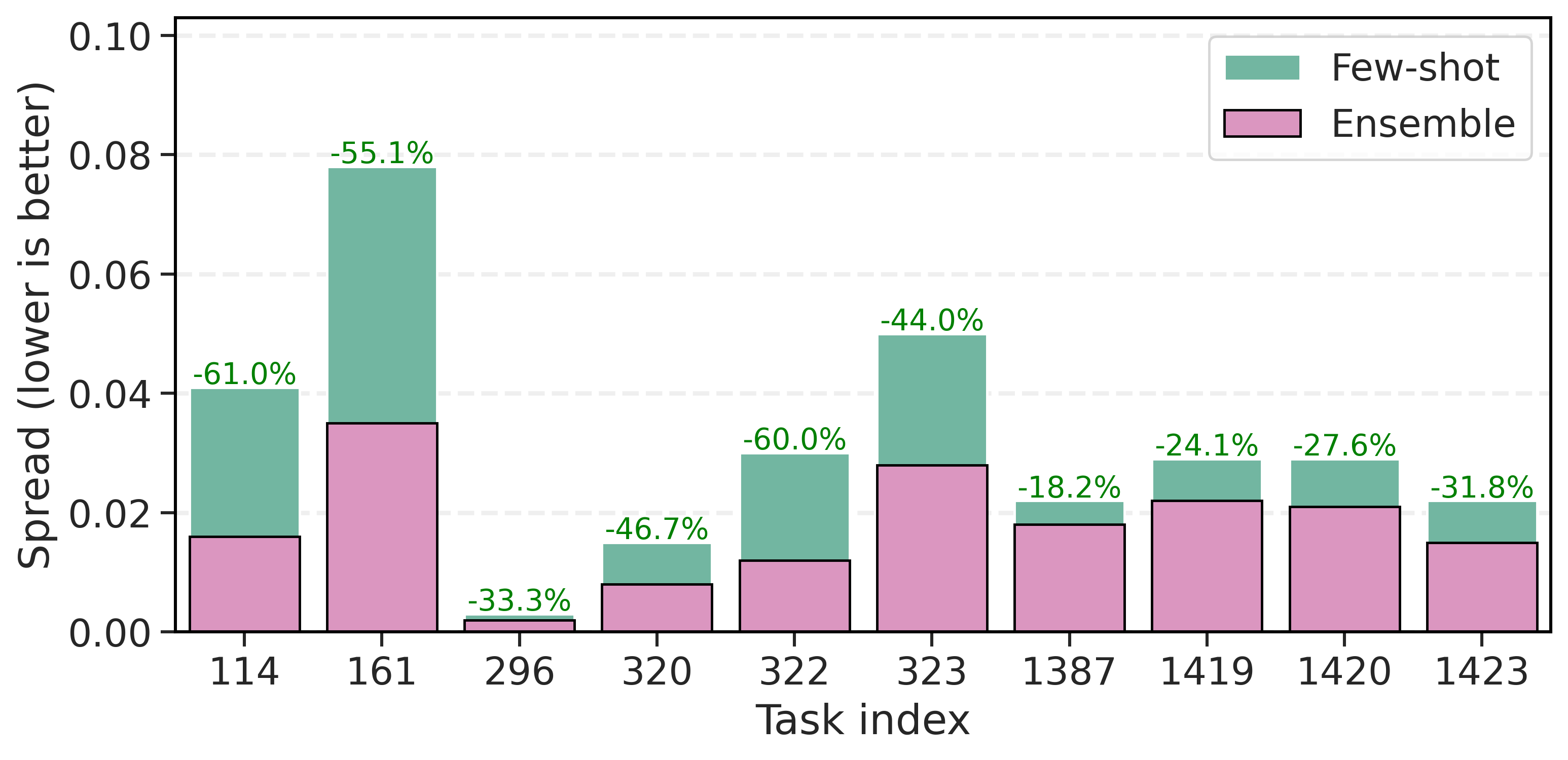}}
    \caption{Spreads for selected tasks from Natural Instructions for two frontier models. Even at this scale, for some tasks spread still might reach 8-10 accuracy points. Using a modified version of Template Ensembles with majority voting instead of probability averaging, we are able to reduce the spread in 19/20 cases, in 9 of which the reduction is at least 44\%.}
    \label{fig:frontier_spread_per_task}
\end{figure*}

\paragraph{Analysis.}
Compositional shift with respect to formats does not affect accuracy and robustness much. We hypothesize this is due to the fact that LoRA with augmentations does not consistently improve robustness even in default scenario considered in \textbf{RQ1}, Section \ref{sec:results_rq1}, and the complexity of compositional shift remains hidden.

Cross-domain transfer is a challenging setup. While it is possible that with another configuration of training data and hyperparameters it might perform better, in our experiments cross-domain fine-tuning with augmentations decreases accuracy below the few-shot baseline.
\medskip

% \begin{takeawayBox}
\textbf{Takeaways:}
\begin{enumerate}[topsep=2pt]
    \item Batch Calibration implicitly assumes a more uniform prior on classes compared to baseline few-shot approach. This inductive bias backfires when the class distribution is skewed.
    \item Cross-domain experiments confirm that high accuracy achieved by LoRA approach substantially relies on training dataset.
\end{enumerate}
% \end{takeawayBox}

\subsection{Greedy Decoding vs. Probability Ranking}
\label{sec:results_rq3}
\textbf{RQ3: How does greedy decoding affect robustness compared to choosing highest-probability answer option?}
To answer \textbf{RQ3}, we run experiments with two inference strategies, described in Section \ref{sec:metrics_and_inference}. Table \ref{tab:ranking_vs_generation} demonstrates, that generation is always less robust to format choice, with Gemma models exhibiting especially large instability. 

Generation approach is widely used in practical applications (chat-bots, API calls), so the problem of format sensitivity can be even more acute there.
\medskip

% \begin{takeawayBox}
\textbf{Takeaway:} greedy decoding exacerbates models' format sensitivity. When possible, opt for probability ranking.
% \end{takeawayBox}

\begin{table*}[h!]
    \centering
    \small 
    \begin{tabular}{llrrr}
    \toprule
    Method & Model & Accuracy $\uparrow$ & \makecell{Std accuracy} $\downarrow$ & Spread $\downarrow$\\
    \midrule
    \multirow[c]{4}{*}{Few-shot} & \textcolor{gray}{Llama 3.1 8B} & \textcolor{gray}{0.563} & \textcolor{gray}{0.052} & \textcolor{gray}{0.161} \\
     & \textcolor{gray}{Qwen 2.5 7B} & \textcolor{gray}{0.605} & \textcolor{gray}{0.058} & \textcolor{gray}{0.190} \\
     & DeepSeek V3 0324 & \textbf{0.741} & 0.015 & 0.045 \\
     & GPT-4.1 & 0.624 & \textbf{0.010} & \textbf{0.032} \\
    \midrule
    \multirow[c]{2}{*}{\makecell{Template Ensembles\\(majority voting)}} & DeepSeek V3 0324 & \textbf{0.742} & 0.009 & 0.028 \\
     & GPT-4.1 & 0.625 & \textbf{0.005} & \textbf{0.018} \\
    \bottomrule
    \end{tabular}
    \caption{Evaluation of frontier models on a subset of 10 out of 52 tasks. For reference, we also include a couple of open-source models. To aggregate accuracy, we take median over formats and average over tasks. Standard deviation and spread are first computed over formats for each task individually and then averaged. \label{tab:frontier}}
\end{table*}

\subsection{Frontier Models \& Format Robustness}
\label{sec:results_rq4_frontier}

We split \textbf{RQ4} into two questions.

\paragraph{RQ4-1: How sensitive are frontier models to format perturbations?} To answer the first part of RQ4, we perform experiments with two frontier models, GPT-4.1 and DeepSeek V3. We evaluate them on a subset of 10 out of 52 tasks due to budget limitations. The results are presented in Table \ref{tab:frontier}. We can see that large closed-source models show much better robustness. While DeepSeek V3 significantly outperforms GPT-4.1 in terms of accuracy, it is more sensitive to format changes.

However, Figure \ref{fig:frontier_spread_per_task} shows that on individual tasks, even frontier models might have spread of 8-10 accuracy points.
 
\paragraph{RQ4-2: What methods can be applied in black-box setting to improve their robustness?} 
To answer the second part of RQ4, we need to consider each method assumptions. Batch Calibration, Sensitivity-Aware Decoding and Template Ensembles require logit access, which is not always available for closed source models. With SFT-based methods the problem is that the user usually has no control over what exact method and hyperparameters are used, even if company provides fine-tuning as a service.

For a broadly applicable approach, we consider an adaptation of Template Ensembles which uses majority voting instead of probability averaging. Figure \ref{fig:frontier_spread_per_task} confirms that this strategy effectively reduces spread, and in Table \ref{tab:frontier} we even see a slight performance improvement, contrary to results of Template Ensembles in Section \ref{sec:results_rq1}. We attribute this to the fact that mode, utilized in majority voting, is substantially more robust to outlier formats than the mean, used in original version.
\medskip
% \begin{takeawayBox}

\textbf{Takeaways:}
\begin{enumerate}[topsep=2pt]
    \item Frontier models are substantially more robust compared to small open-source models, suggesting that scaling improves robustness.
    \item Occasionally, there are still cases where spread can reach 8-10 accuracy points. To deal with them, Template Ensembles with majority voting might be used.
\end{enumerate}
% \end{takeawayBox}
\FloatBarrier
\section{Conclusion}

To the best of our knowledge, we have conducted the first comprehensive comparison of existing prompt sensitivity mitigation methods across multiple model families, sizes and distribution shifts.

We provide actionable insights for practitioners. For example, excessive fragility of calibration-based methods to class imbalance underscores the consequences of implicit assumptions, and hints that a reliable estimate of prior is important.
Meanwhile, ineffectiveness of light supervised finetuning with augmentations at improving robustness suggests that more research is needed to develop a strong baseline in this paradigm. 
Finally, our experiments on frontier models confirm that scale is positively correlated with robustness. Still, on some tasks even large models exhibit 8-10 accuracy point differences solely due to format changes. Version of Template Ensembles with majority voting  helps to mitigate this instability.

We also release our code to facilitate research aimed to address the problem of format sensitivity.
\section*{Limitations}

Our study provides deep insights into classification and multiple-choice tasks, leaving more complex settings like text generation or multi-step reasoning out of the scope. Nonetheless, we underline that models exhibits substantial instability even in this simple setting.

Some of considered robustness methods are harder to apply to frontier models. Batch Calibration, Sensitivity-Aware decoding and Template Ensembles require access to logits, which are sometimes unavailable. Finetuning approaches might be expensive at large scale. Additionally, when using finetuning API, users usually have limited control over the finetuning procedure.

%While we demonstrate clear trends with models up to 9B parameters, testing these methods on larger language models would help understand how format sensitivity evolves with scale. 
%It would be particularly interesting to explore whether the effectiveness of methods like Batch Calibration continues to improve with model size, given our observation that larger models generally show better format robustness.

\section*{Ethics}

The models and datasets used in this study are publicly available for research purposes, with licenses detailed in Appendix \ref{sec:licenses}. All experiments were performed on NVIDIA A100 80GB GPUs. Each fine-tuning or evaluation run was conducted on a single GPU, and took between 1 and 24 hours, depending on the model size and method's efficiency. To optimize computation, we utilized up to 46 GPUs in parallel. In total, the experiments took approximately 15,000 GPU-hours. Our PyTorch/Hugging Face code will be released alongside the paper, and we expect no direct social or ethical concerns arising from this work.

\paragraph{Use of AI Assistants} We utilize Grammarly to enhance and proofread the text of this paper, correcting grammatical, spelling, and stylistic errors, as well as rephrasing sentences. Consequently, certain sections of our publication may be identified as AI-generated, AI-edited, or a combination of human and AI contributions.
We also used DeepSeek V3, Claude Sonnet 3.5 and ChatGPT to improve text fluency and implement some of the code for results visualization.

\FloatBarrier
\bibliography{anthology, custom}
\appendix
\section{Consistency loss}
\label{sec:appendix_consistency_loss}

\begin{figure*}
    \centering
    \includegraphics[width=\linewidth]{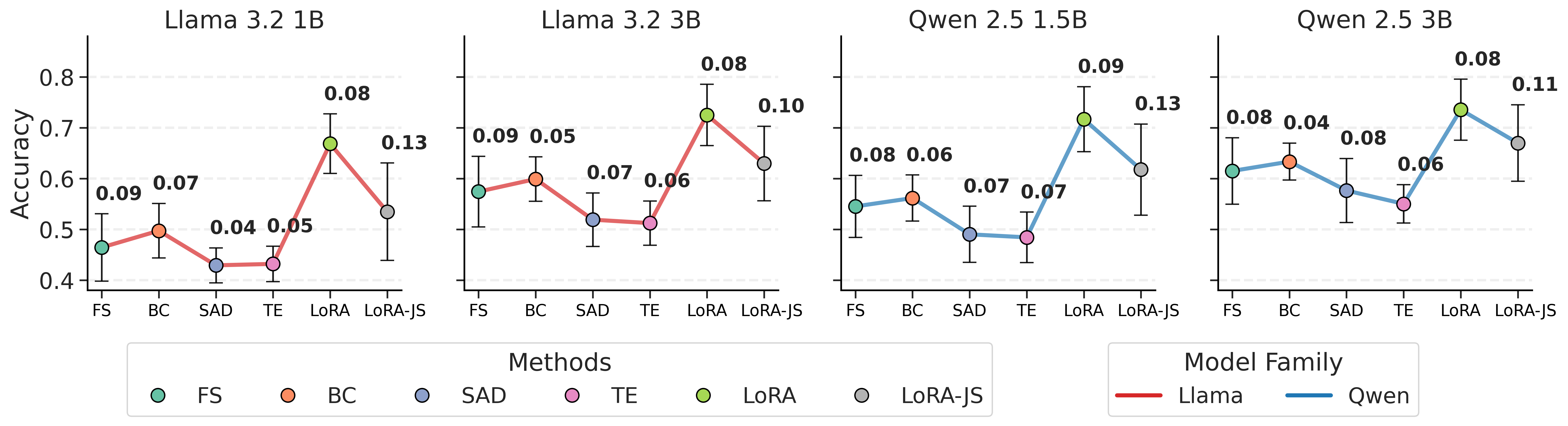}
    \caption{Comparing LoRA with consistency loss (LoRA-JS) with other methods in terms of their effect on accuracy and standard deviation over prompt formats. To aggregate accuracy, we first compute median accuracy over formats for each task, and then average over 52 tasks. Error bars are $2\;\times$ (standard deviation over formats, averaged across tasks).}
    \label{fig:consistency}
\end{figure*}

Figure \ref{fig:consistency} provides some results for LoRA with consistency loss. 
Compared to LoRA with format augmentations, it shows higher spread and lower accuracy for all models.

\section{Extended Description of Methods}
\label{sec:full_methods_description}

In this section, we provide a more detailed description of the methods used in the paper.

\paragraph{Batch Calibration (BC).} 

Batch Calibration \cite{zhou2024batchcalibration} is a post-hoc approach, which calibrates model prediction with the estimate of contextual bias term $p(y \,| \, C)$. The contextual bias for each class $p(y = y_j | C)$ is estimated from a batch of $B$ samples by marginalizing the output scores over all samples within the batch. The calibrated probabilities $\hat{y}_i$
are derived by shifting the log-probability $ \log p(y | x_i, C)$ by the corresponding estimated mean of each class:
\begin{gather}
    \forall y_{j} \in Y\colon \\
    \nonumber \overline{\log p(y | C)_{j}} = \frac{1}{B} \sum \limits_{i=1}^{B} \log p(y = y_{j} | x^{(i)}, C),  \\
    \nonumber \hat{y}^{(i)} = \argmax \limits_{y \in Y} \left( \log p(y | x^{(i)}, C) - \overline{\log p(y | C)} \right).
\end{gather}

The method was originally designed exclusively for classification, and is inapplicable to other tasks.

\paragraph{Template Ensembles (TE).}

Template Ensembles \cite{voronov-etal-2024-mind} average the predictions probability across various prompt formats $f_i$ and select the class with the largest probability. 

\begin{equation}
    \hat{y} = \argmax_{y \in Y} \frac{1}{N} \sum_{i=1}^N p(y \, | \, x, f_i)
\end{equation}

The main drawback of this method is the need to run the model $N$ times, where $N$ is the ensemble size.

\paragraph{Sensitivity-aware decoding (SAD).}

Sensitivity-aware decoding \cite{lu-etal-2024-prompts} assesses model sensitivity to input data by evaluating the variance of $N$ predictions over modified inputs, using synthetic perturbations based on real data. The sensitivity value $s$ is then used as a penalty in the greedy decoding process: 
\begin{equation}
    \hat{y} = \argmax \limits_{y \in V} \left[ \alpha P(y|x) - (1 - \alpha)s \right],
\end{equation}
where $P(y|x)$ is the probability of an output $y$ given $x$, $\alpha$ is the reweighting hyperparameter and $V$ is vocabulary -- the set of all tokens. 
In this paper we use a simplified version of sensitivity-aware decoding, using random token substitutions as a perturbation. This approach reduces model variance but also requires $N$ times more runs.

\paragraph{LoRA with augmentations (LoRA).}

We fine-tune an instruction-finetuned model $M$ on a small dataset $D_\text{format}$ containing augmented samples. Here augmentation refers to changing the formatting while maintaining the same content. To build $D_\text{format}$, we select a subset of samples from a generic instruction-following dataset $D_\text{source}$ and insert $K$  augmented versions for each sample.

Parameter-efficient finetuning on $D_\text{format}$ is conducted using a standard language modeling cross-entropy loss. Loss is only computed on answer tokens, while the prefix tokens are masked.

\paragraph{LoRA with consistency loss (LoRA-JS).}

One way to enforce consistent predictions is to use an auxiliary loss during fine-tuning. To test this approach, we reproduce \emph{prompt perturbation consistency learning} \cite{qiang2024consistency}.

The training objective contains supervised cross-entropy losses for pairs of augmented examples $x_1, x_2$ that share the same target label $y$,  along with the consistency loss, based on Jensen-Shannon divergence.
Formally, the overall loss function is defined as:
\begin{align} 
    \mathcal{L} = \text{CE}(\hat{y}_1, y) + \text{CE}(\hat{y}_2, y) + \beta \text{JS}(\hat{\bar{y}}_1 || \hat{\bar{y}}_2),
    % \mathcal{L}_m = \text{CE}(\hat{y}_m, y), \\
    % \mathcal{L}_n = \text{CE}(\hat{y}_n, y), \\
    % \mathcal{L}_{mn} = \text{JS}(\hat{\bar{y}}_m || \hat{\bar{y}}_n)
\end{align}

where $\text{CE}$ denotes the cross-entropy loss, $\beta$ is the coefficient controlling the contribution of the consistency loss, $\text{JS}$ is Jensen-Shannon divergence, $\hat{y_1}, \hat{y_2}$ are the response token probability distributions, and $\hat{\bar{y}}_1$ and $\hat{\bar{y}}_2$ are corresponding distributions averaged over response length.
\section{Method hyperparameters}
\label{sec:hyperparameters}

All LoRA models were trained with default hyperparameters, given in Table \ref{tab:training_hyperparameters}.

\begin{table}[]
    \centering
    \small
    \begin{tabular}{lc}\toprule
    Learning rate & 2e-4 \\
    LoRA $\alpha$ & 16 \\
    LoRA Rank & 16 \\
    Amount of epochs & 1 \\
    Batch size & 64 \\
    Weight decay & 0.01 \\
    % Consistency loss $\beta$ & 50 \\
    \bottomrule
    \end{tabular}
    \caption{LoRA training hyperparameters.}
    \label{tab:training_hyperparameters}
\end{table}

\paragraph{LoRA fine-tuning data.} \label{par:lora_ft_details} 
For experiments in Section \ref{sec:results_rq1} we construct $D_\text{format}$ from our subset of Natural Instructions benchmark, choosing up to $1000$ samples per task, disjoint with the test samples chosen before.
For cross-domain experiments in \ref{sec:results_rq2_shifts} we use a custom fine-tuning dataset built from a subsample of the Open Hermes 2.5 dataset \cite{OpenHermes25}. Open Hermes 2.5 contains synthetically generated tasks in the form of prompts for LLMs, covering various task types. 
Our research primarily focuses on classification tasks and multiple-choice questions. Although the dataset includes many such tasks, only a small portion has clearly defined labels. 
To find such examples, we selected those where the GPT response length does not exceed 20 symbols, as suitable tasks typically feature simple and concise answers.
This threshold was determined empirically.
Resulting dataset has approximately $50$k samples.

\paragraph{Other methods' hyperparameters.} In experiments with Template Ensembles and Sensitivity-aware decoding, we used ensembles of size 5, following \cite{voronov-etal-2024-mind}. 

The $\alpha$ parameter in Sensitivity-aware decoding was set to $0.7$ based on Section A.7 in \cite{lu-etal-2024-prompts}. To create synthetic inputs for sensitivity estimation, we replaced 15\% of the original tokens with random tokens from the entire vocabulary.

\begin{table}[h]
    \centering
    \small
    \begin{tabular}{cl}\toprule    
        Format 1 & \texttt{Question: \{\} Answer: \{\}} \\ 
        Format 2 & \texttt{Question:: \{\} Answer:: \{\}} \\ 
        Format 3 & \texttt{QUESTION\textbackslash n\{\}\textbackslash nANSWER\textbackslash n\{\}} \\
        Format 4 & \texttt{question -- \{\} answer -- \{\}} \\
        \bottomrule
    \end{tabular}
    \caption{Some of augmentations used during LoRA finetuning. \label{tab:aug_examples}}
\end{table}

\section{Task selection}
\label{sec:selection}

We selected tasks from Super-Natural Instructions following the criteria outlined in \cite{sclar2024quantifyinglanguagemodelssensitivity}.

A total of 52 evaluation tasks were selected from Super-Natural Instructions using several heuristics. First, datasets were required to contain at least 1000 samples. Then, tasks with long instructions (over 3000 characters) and inputs (over 2000 characters) were excluded to ensure scalability. Additionally, tasks with a predicted accuracy of 0\% for LLaMA-2-7B 1-shot were removed, and no more than 4 tasks from the same dataset were included. Furthermore, socially significant tasks and formats for them were added if they were missing.

The selected tasks were the following 52: \texttt{task050, task065, task069, task070, task114, task133, task155, task158, task161, task162, task163, task213, task214, task220, task279, task280, task286, task296, task297, task316, task317, task319, task320, task322, task323, task325, task326, task327, task328, task335, task337, task385, task580, task607, task608, task609, task904, task905, task1186, task1283, task1284, task1297, task1347, task1387, task1419, task1420, task1421, task1423, task1502, task1612, task1678, task1724}.

\texttt{task190} fits all previous requirements, but was excluded due to labeling errors.

For evaluation of frontier models, we selected \texttt{task114, task161, task296, task320, task322, task323, task1387, task1419, task1420, task1423}, since they cover math, text compherension, simple tasks like counting the number of words with a given letter and socially significant topics like detecting racial stereotypes.
\section{Dependence between spread and format complexity}

\begin{figure}
    \centering
    \includegraphics[width=\linewidth]{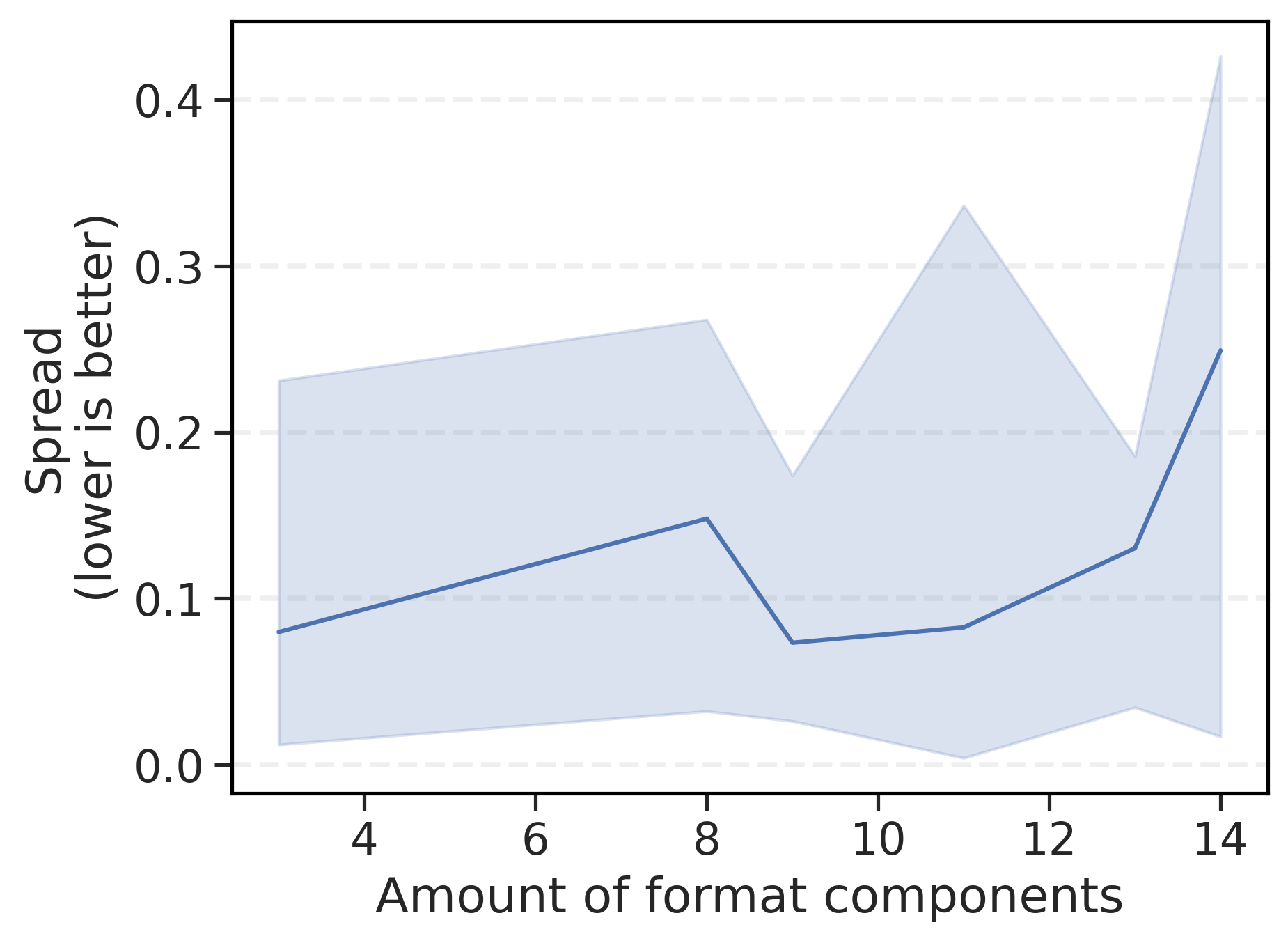}
    \caption{Empirical dependency between spread and amount of components in format. 90\% confidence interval is based on percentiles.}
    \label{fig:spread_vs_n_format_components}
\end{figure}

Figure \ref{fig:spread_vs_n_format_components} shows the relation between spread and format complexity, measured as the amount of prompt components. While the dependence is quite noisy, maximal spread values occur at formats of maximal length.

\section{Use of scientific artifacts.}
\label{sec:licenses}
\begin{table}[hb!]
    \centering
    \small
    \begin{tabular}{cc}\toprule
        Artifact & License \\\midrule
        Natural Instructions & Apache 2.0 \\
        Open Hermes 2.5 & CC-BY-NC \\
        Llama 3.1 8B & \makecell{Llama 3.1 Community\\License Agreement} \\
        Llama 3.2 (1B, 3B) & \makecell{Llama 3.2 Community\\License Agreement} \\
        Gemma 2 (2B, 9B) & Apache License 2.0 \\
        Qwen2.5 (1.5B, 3B, 7B) & Apache License 2.0 \\ \bottomrule
    \end{tabular}
    \caption{Scientific artifacts used in this paper and their licenses.}
    \label{tab:artifacts}
\end{table}

In Table \ref{tab:artifacts} we list the artifacts used in this paper along with their licenses. To the best of our knowledge, using these artifacts for research purposes is consistent with their intended use.
\FloatBarrier
\section{Complete list of format components.}
\label{sec:all_format_components}

Complete list of format components is given in Table \ref{tab:all_format_components}.

\begin{table*}[]
    \centering
    \small
    \begin{tabular}{cc}
        \toprule
        Modification & Values \\
        \midrule
        Descriptor transformation & \texttt{\makecell{lambda x: x.title(), lambda x: x.upper(),\\ lambda x: x.lower(), lambda x: x}} \\\midrule
        Separator & \texttt{\makecell{'', '::: ', ':: ', ': ', ' \textbackslash n\textbackslash t', '\textbackslash n ',\\ ' : ', ' - ' , ' ', '\textbackslash n ', '\textbackslash n\textbackslash t', ':',\\ '::', '- ', '\textbackslash t'}} \\\midrule
        Space & \texttt{\makecell{'', ' ', '\textbackslash n', ' \textbackslash n', ' -- ', ' ', '; \textbackslash n',\\ ' || ', ' <sep> ', ' -- ', ', ', ' \textbackslash n ',\\ ' , ', '\textbackslash n ', '. ', ' , '}} \\\midrule
        Text \& option separator & \texttt{'', ' ', ' ', '\textbackslash t'} \\\midrule
        Option item style & \texttt{\makecell{1, 2, ...; A, B, ...; a, b, ...; \\I, II, ...; i, ii, ...}} \\\midrule
        Option item wrapper	& (\{\}); \{\}.; \{\}); \{\} ); [\{\}]]; <\{\}> \\
        \bottomrule
    \end{tabular}
    \caption{Descriptor transformation correspond to Python commands making first character upper case (title), all letters upper case (uppercase), all letters lower case (lowercase) or keeping input as is. Option item style includes Arabic numerals, uppercase and lowercase Latin letters and uppercase and lowercase Roman numerals. For option item wrapper, $\{\}$ is used as a placeholder for option item (e.g. `a' or `1').}
    \label{tab:all_format_components}
\end{table*}

\newenvironment{allintypewriter}{\ttfamily}{\par}
\section{Example of input prompt for frontier models}

\begin{figure*}
    \centering
    \begin{takeawayBox}[width=\textwidth]
    \begin{allintypewriter}
    System: In this task, you need to answer 'Yes' if the given word is the longest word (in terms of number of letters) in the given sentence, else answer 'No'. 
    Note that there could be multiple longest words in a sentence as they can have the same length that is the largest across all words in that sentence. 
    PAY ATTENTION TO THE OUTPUT FORMAT -- ONLY OUTPUT THE ANSWER WITHOUT ANY OTHER TEXT, LIKE IN EXAMPLES.\\
    \\
    User: Sentence\\
     'woman sitting on a chair holding three teddy bears'. Is 'a' the longest word in the sentence?\\
     Answer\\
     No\\
    \\
    Sentence\\
     'a large green plant with leaves and spiky flowers'. Is 'flowers' the longest word in the sentence?\\
     Answer\\
     Yes\\
    \\
    Sentence\\
     'a long white airplane covered with a lot pastel hears on it'. Is 'covered' the longest word in the sentence?\\
     Answer
    \end{allintypewriter}
    \end{takeawayBox}
    \caption{Example of input prompt used for GPT-4.1 and DeepSeek V3 0324. \label{fig:example_prompt}}
\end{figure*}

Example of input prompt used for frontier models in presented on Figure \ref{fig:example_prompt}.

\appendix

\end{document}